# MOANA: An Online Learned Adaptive Appearance Model for Robust Multiple Object Tracking in 3D

**Zheng Tang, Student Member, and Jenq-Neng Hwang, Fellow, IEEE**
Department of Electrical and Computer Engineering, University of Washington, Seattle, WA 98195 USA
Corresponding author: Zheng Tang (e-mail: zhtang@uw.edu).

**ABSTRACT** Multiple object tracking has been a challenging field, mainly due to noisy detection sets and identity switch caused by occlusion and similar appearance among nearby targets. Previous works rely on appearance models built on individual or several selected frames for the comparison of features, but they cannot encode long-term appearance changes caused by pose, viewing angle and lighting conditions. In this work, we propose an adaptive model that learns online a relatively long-term appearance change of each target. The proposed model is compatible with any feature of fixed dimension or their combination, whose learning rates are dynamically controlled by adaptive update and spatial weighting schemes. To handle occlusion and nearby objects sharing similar appearance, we also design cross-matching and re-identification schemes based on the application of the proposed adaptive appearance models. Additionally, the 3D geometry information is effectively incorporated in our formulation for data association. The proposed method outperforms all the state-of-the-art on the *MOTChallenge* 3D benchmark and achieves real-time computation with only a standard desktop CPU. It has also shown superior performance over the state-of-the-art on the 2D benchmark of *MOTChallenge*.

## I. INTRODUCTION

In recent years, the unprecedented explosion in the availability of and access to image big data has contributed to the rapid development of computer vision algorithms. Especially, the performance of object detectors has been improved dramatically in the last two decades. As a result, more and more attention has been drawn to the study of tracking by detection, i.e., the observations of objects are generated by object detection that may contain some errors. With the detected observations in each video frame as input, the goal of multiple object tracking (MOT) is to recover the trajectories of all targets in a video sequence. MOT is of high significance to many useful applications in computer vision and robotics, e.g., security surveillance, autonomous driving, etc. Though MOT has seen considerable progress in recent years because of improved appearance models and optimization schemes, the status quo is still far from matching human performance. The major challenges include noise in object detection, appearance change, and identity switch caused by object occlusion and similar appearance between objects in pair/group. The problems can be mitigated when the camera projection matrix is available, which can convert the tracking space into 3D. Thus, the depth information can be effectively utilized, whereas the prediction of object movement and scale can be more reliable.

Most of the state-of-the-art methods focus on data association techniques. The majority of them are offline algorithms, e.g., [1], [2], [3], in which observations of objects are grouped into tracklets based on spatio-temporal continuity. In data association, besides motion patterns and social force models, appearance models have also been widely used as an important cue to keep the identities of targets. Traditionally, appearance models based on raw pixel template representation [4], [5], fusion of color/texture/edge features [6], or color/texture/edge histograms [3, [7], [8], [9], [10] are adopted for their simplicity. Nevertheless, these models are only built on individual or several selected frames, which could not encode long-term appearance change along each trajectory. Thus, they may fail when there is change of lighting condition, viewing angle or object pose. Other researchers also introduce methods based on *random forest* algorithms [11], [12] or take advantage of deep learning features [13] to improve the robustness of appearance modeling, but the computation complexity significantly increases and massive training samples are required.

Inspired by adaptive background modeling in change detection [14], [15], [16], [17], we propose an adaptive appearance model that can learn the long-term change of object appearance online. As partially described in [18], [19], the proposed framework, termed MOANA which is short for "Modeling of Object Appearance by Normalized Adaptation," models the appearance of each target as a normalized matrix with an array of observed feature vectors at each cell. MOANA is compatible with any feature of fixed dimension or their combination. To update the model, the learning rates are controlled by the similarity with previous features and spatial weighting. When an object is partially occluded by or spatially close to others, a cross-matching module is employed to avoid identity switch based on the proposed appearance model. For objects that are seriously occluded or failed to be detected (false negatives) for a few frames, we design a re-identification scheme to recover their trajectories. 3D geometry information is also leveraged in our formulation of data association. Experiments are conducted on the test and training sets of the *MOTChallenge* 3D benchmark [20]. We are ranked on top of the benchmark in terms of the multiple object tracking accuracy (MOTA) [21]. [1] Our proposed method has also shown superior performance over the state-of-the-art in the *MOTChallenge* 2D benchmark [20].

The major contribution of this work is three-fold. (1) We propose an adaptive model that can encode long-term appearance change for robust object tracking, which is inspired from adaptive background modeling in change detection. (2) Cross-matching and re-identification schemes are designed to overcome occlusion and ambiguity among neighboring objects, which incorporate both the adaptive appearance model and 3D geometry information. (3) The proposed framework achieves superior performance over the state-of-the-art on the *MOTChallenge* benchmark collection.

The rest of this paper is organized as follows. In Section II, related works of our approach are reviewed in detail. The system overview and description of each algorithmic component are covered in Section III. The implementation details and evaluation of our method on the *MOTChallenge* benchmarks are presented in Section IV. Finally, we draw the conclusion in Section V.

## II. RELATED WORK

### A. MULTIPLE OBJECT TRACKING BY DETECTION

One of the traditional approaches to MOT is to predict the states, i.e., location and size, of tracked targets based on Bayesian inference methods, e.g., Kalman filter or particle filter [4], [8], [10]. These methods usually can achieve acceptable performance in short term, however, they tend to fail when objects are interacting with each other, i.e., under occlusion and/or movement in groups.

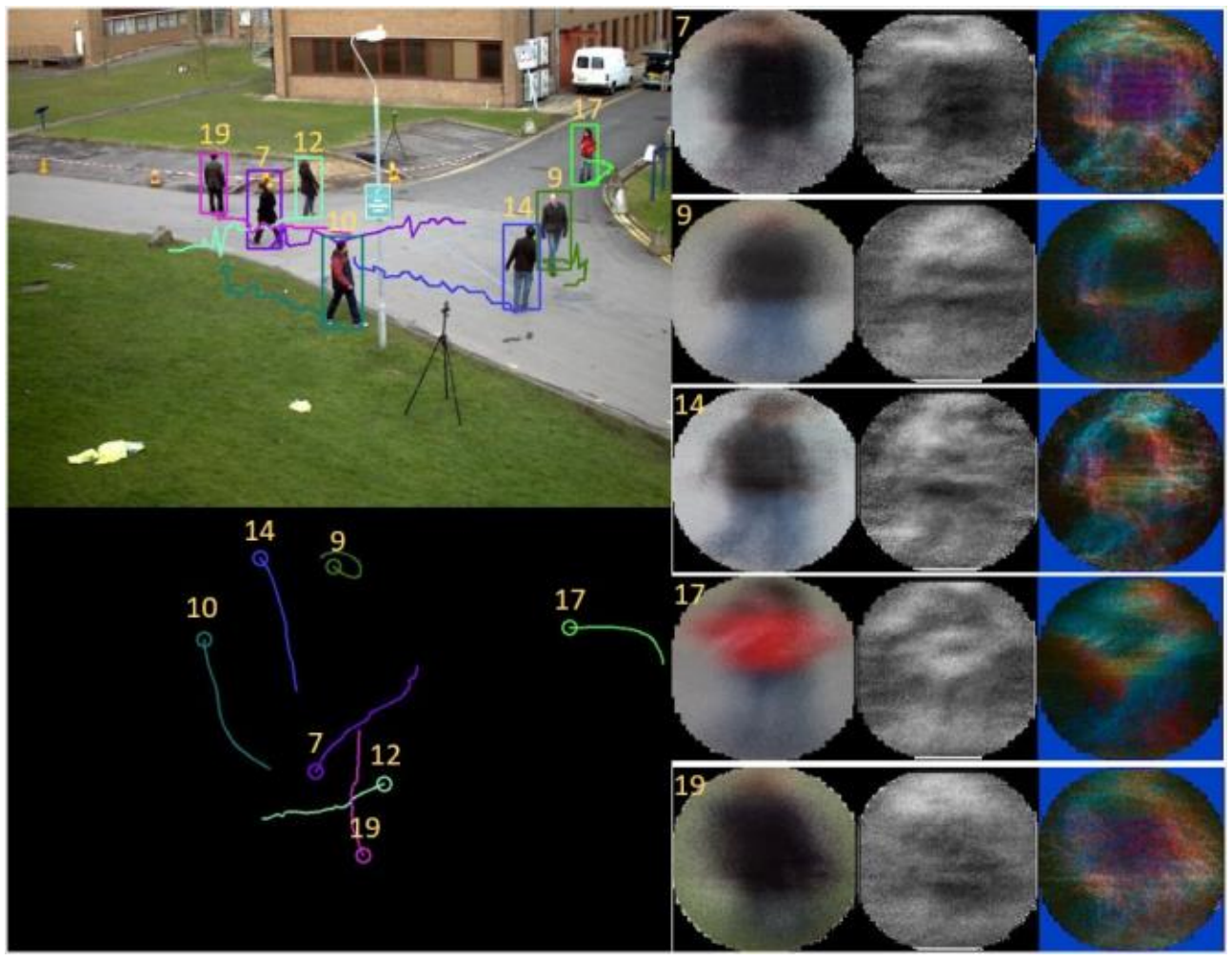

**FIGURE 1. Multiple object tracking in 2D (*top-left*), the back projection to 3D in top view (*bottom-left*) and the visualization of the averaged adaptive appearance models learned online in RGB space, LBP space, and gradient space (*right*).**

Many recent works formulate MOT as a data association problem. Leal-Taixé et al. [1] propose to formulate data association by social force and grouping behavior. The *probability hypothesis density* (PHD) filter [2] is introduced in the formulation of multi-target state estimation for offline decision on data association. Wen et al. [3] uses a space-time-view hyper-graph to encode higher-order constraints in 3D. More recently, some researchers apply deep learning architectures like *recurrent neural networks* (RNNs) to the modeling of nonlinear behaviors in data association [13], [22].

Relatively little attention has been given to the development of discriminative appearance models for MOT. Methods like [4], [5], [6], [3], [7], [8], [9], [10] employ raw pixel template representation or fusion of traditional image features from a single frame to model the object appearance. The histogram representation is improved by Chu et al. [7], who build multiple spatially weighted kernel histograms with binding constraints for each target to overcome partial occlusion. Similarly, Yang and Nevatia [23] introduce *discriminative part-based appearance models* (DPAMs), which uses a human part model to extract the discriminative features from unoccluded object area. However, all the mentioned appearance models are highly sensitive to the quality of the selected frame(s), which may fail occasionally due to illumination or other conditions. Besides, their similarity measurements either not or only implicitly encode the spatial distribution of appearance features. On the other hand, Kuo et al. [24] present *online learned discriminative appearance models* (OLDAMs) to learn the discriminative features from training samples collected online with some spatio-temporal constraints. In [25], the *conditional random field* (CRF) model is exploited to combine OLDAMs with non-linear motion patterns. Though the affinity measurement can be learned online for [24] and [25], they still only consider features extracted from single frame such as RGB color histogram and *histogram of oriented gradients* (HOG). These features can lead to tracking errors after objects being occluded for long time. There are other attempts to adapt the classifier to the changing appearance of each target by using variants of *random forests* [11], [12] and boosting [26]. Some also propose to apply deep learning features generated from the *convolutional neural networks* (CNNs) [13] to improve tracking performance. However, because of the increased complexity of these methods and the potentially large number of targets, the computation requirement becomes a major challenge. Moreover, these methods require massive training samples to achieve robust performance. Recently, Ma et al. [27] introduce a novel Dirichlet-process-based statistical model to describe the underlying distribution of non-Gaussian image features. Based on their feature modeling, the performance on various tasks is significantly improved compared to the state-of-the-art, which also verifies the benefit of appearance modeling for pattern recognition problems.

[1]Available at https://motchallenge.net/results/3D_MOT_2015/

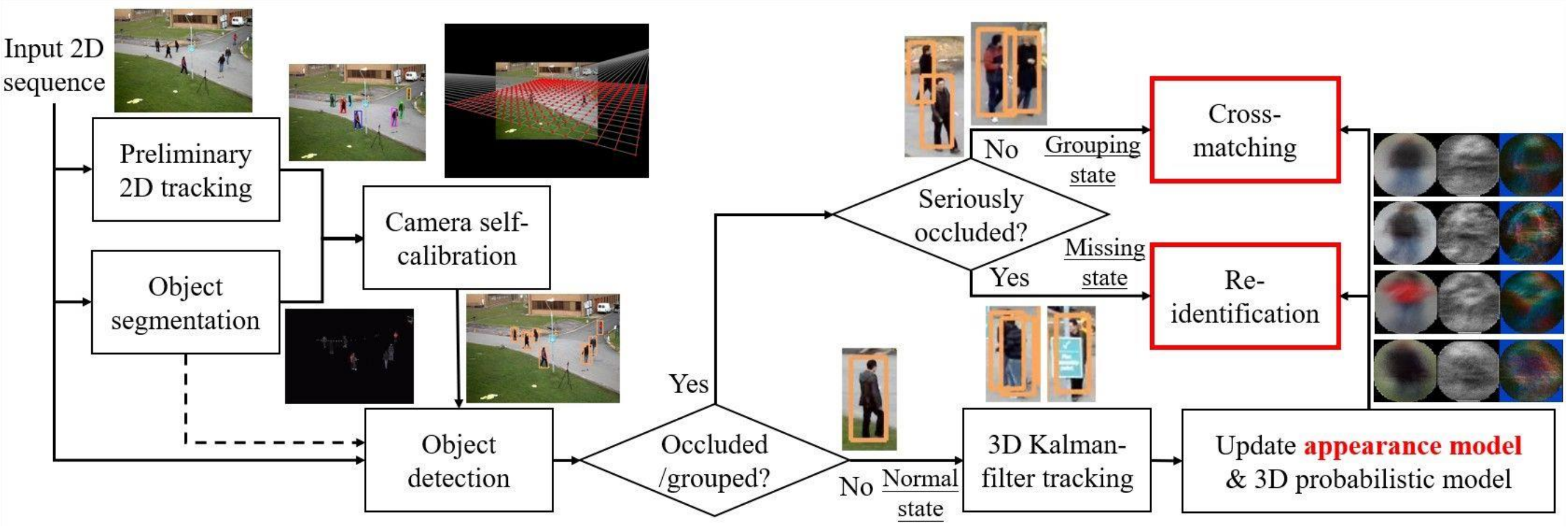

**FIGURE 2.** Flow diagram of the proposed MOT framework based on adaptive modeling of object appearance.

### B. BACKGROUND MODELING IN CHANGE DETECTION

Background modeling is a key element of modern change detection algorithms. Barnich et al. [14] introduce the *visual background extractor* (ViBe) that builds the background model with a set of observed values in the past at each pixel location. The *pixel-based adaptive segmenter* (PBAS) [15] improves the pixel-based background modeling scheme by applying a random observation replacement policy. The *self-balanced sensitivity segmenter* (SuBSENSE), proposed by St-Charles et al. [16], [17], further improves the update scheme using pixel-level feedback loops that dynamically adjust the internal configuration parameters. To the best of our knowledge, our work is the first to extend adaptive modeling and random update scheme in change detection to support robust object tracking. We also design cross-matching and re-identification schemes to resolve ambiguity among objects using the adaptive appearance models.

## III. METHODOLOGY

The overview flow diagram of our proposed framework is shown in Fig. 2. We first exploit the output of preliminary 2D human tracking and foreground segmentation [10] for camera self-calibration [28], [29]. The observations of objects can be located by object detection or with the assistance of segmentation. When a target is not occluded by or grouped with other object(s), it is associated with available observation(s) based on an efficient 3D Kalman-filter-based strategy. The proposed appearance models and a probabilistic model of 3D object properties are learned online. When an observation is grouped with others, the cross-matching module is enabled to associate nearby targets based on the unoccluded area of appearance models. On the other hand, when an object is seriously occluded or missing, his/her appearance model is temporarily stored and used for re-identification. The detailed formulation and the role of each component are illustrated as follows.

### A. FORMULATION OF DATA ASSOCIATION

Before introducing the proposed adaptive appearance model for object tracking, we first define the formulation of MOT as a data association problem in time and space. We aim to recover the trajectories $T$ of all targets within the 3D scene, which are defined as

$$T = \{T_i : i = 1,2,\dots,|T|\}, \quad (1)$$

where each $T_i$ is equivalent to an object identity.

The basic units of MOT are the candidate observations of objects, noted $O$, derived from object detection or with the assistance of foreground segmentation, defined as

$$O = \{O_j \sim (g_j, f_j, q_j, t_j) : j = 1,2,\dots,|O|\}, \quad (2)$$

in which $g_j$ is the 3D geometry information, $f_j$ is the extracted appearance feature, $q_j$ is the foreground mask within the object region, and $t_j$ is the time stamp. They will be illustrated in detail in the following subsections.

The goal of MOT is to solve the following objective from an input video sequence

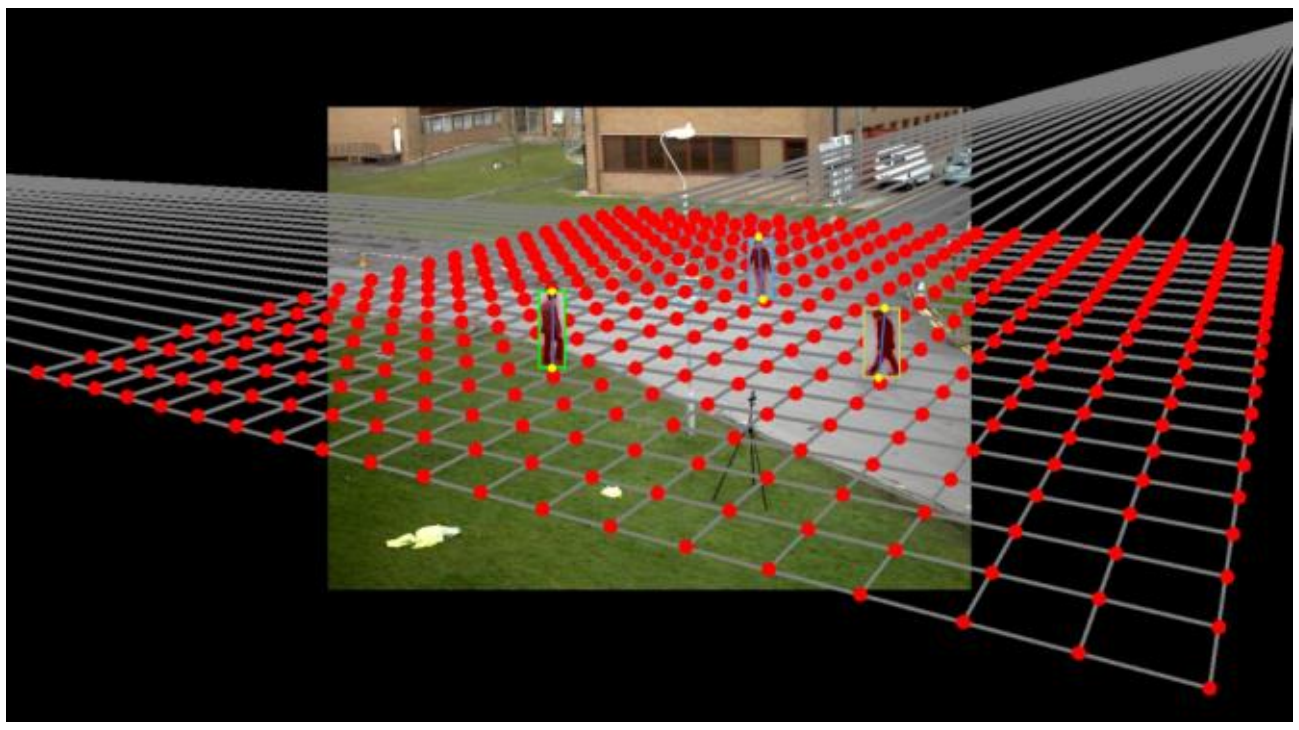

**FIGURE 3.** Projected 3D grid on the ground plane generated by camera self-calibration with the extracted head and foot points highlighted.

$$T_i \leftarrow O_j, \forall i, \forall j, \quad (3)$$

which represents the assignment of every observation to a corresponding object identity. The false positives are all assigned to $T_\infty$.

When the camera parameters are unavailable, we first process a short period of the video sequence by preliminary 2D tracking and foreground segmentation [10]. Each human object is modeled as a pole perpendicular to the ground plane, whose endpoints are located based on the orientation of the foreground blob, from which we can compute the horizon line and vanishing points in the scene for camera self-calibration [28], [29]. An example of the estimated 3D ground plane from camera self-calibration is shown in Fig. 3. Then, we process the sequence from the beginning with each object observation back projected to 3D space. The geometry information of each $O_j$, noted $g_j$, is represented by six aspects.

$$g_j \sim (b_j, P_j, D_j, V_j, W_j, H_j), \quad (4)$$

where $b_j \in \mathbb{R}^4$ denotes the 2D bounding box represented in terms of centroid coordinates and size, $P_j \in \mathbb{R}^2$ denotes the back projected foot point coordinates on the 3D ground plane, i.e., the X-Y plane, $D_j$ denotes the 3D depth of $P_j$, $V_j \in \mathbb{R}^2$ denotes the 3D velocity of $P_j$ on the X-Y plane, and $W_j$ and $H_j$ are the width and height of the 3D bounding box, respectively.

An observation $O_j$ is deemed to be under occlusion or grouped with other object(s) if $b_j$ overlaps with other(s) or the 3D distance of their foot points is smaller than a threshold $\tau_P$. Otherwise, $O_j$ is associated with a $T_i$ based on an efficient 3D Kalman-filter-based approach. The state vector of the Kalman filter has six dimensions, corresponding to $P_j$, $V_j$, $W_j$ and $H_j$, whose prediction and update are similar to the 2D scenario [8].

The Kalman prediction of a target $T_i$ is regarded as a predicted observation, noted $\widehat{O}_i$ . An observation $O_j$ is associated with $T_i$ based on the following rule

$$T_i \leftarrow O_j, \text{ if } \frac{\|\widehat{P}_i - P_j\|_2}{w_j^D} < \tau_P, \quad (5)$$

which means that the predicted 3D foot point $\widehat{P}_i$ of $T_i$ is within a short Euclidean distance of $O_j$. The term $w_j^D$ is proportional to the depth of $O_j$, defined by

$$w_j^D = D_j \cdot \eta_D + c_D, \quad (6)$$

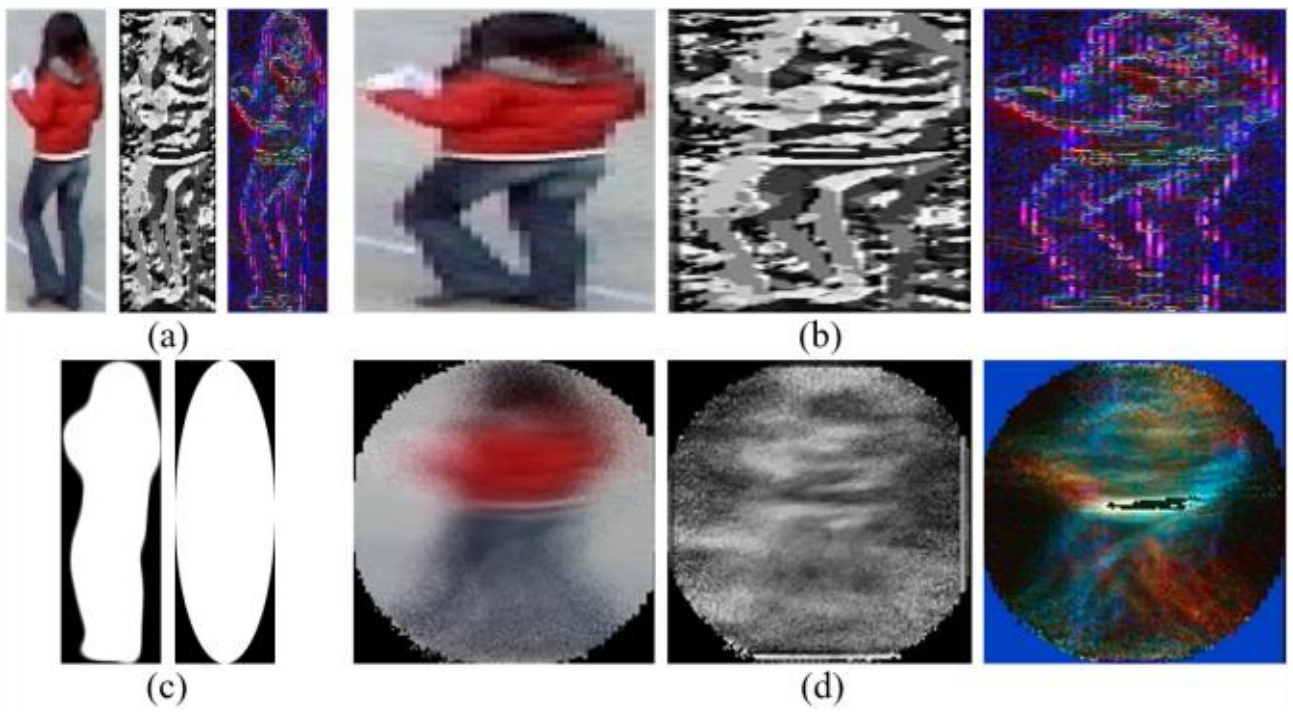

**FIGURE 4.** An example of the construction and update of MOANA. (a) The RGB image for color representation, the LBP image for texture representation and the gradient image for edge representation. (b) Feature maps normalized to $w \times h$. (c) The foreground masks used to indicate visible object area to be updated. When the segmentation results are not available, a maximum-ellipse mask is used. (d) The visualization of the averaged feature components in the adaptive appearance model.

where $\eta_D$ is a constant step size and the addition of a constant $c_D$ is to avoid division-by-zero error. The intention of the division by $w_j^D$ in Equ. (5) is to compensate for the ambiguity in 3D measurement of distant objects, whose estimated 3D foot points are highly sensitive to small errors in object detection and/or foreground segmentation.

When tracking under the mode of Kalman filtering, we also build a probabilistic model of 3D object properties online. The probabilistic model has four dimensions, corresponding to $V_j$, $W_j$ and $H_j$. A four-dimension probabilistic model is used to actively learn the normal distribution of each 3D property. False positives of object observations are removed from the list of candidates for association based on the three-sigma rule of thumb in normal distribution.

### B. ADAPTIVE MODELING OF OBJECT APPEARANCE

Even though 3D Kalman-filter-based tracking can generate more reliable tracklets compared to 2D tracking, it still cannot overcome the problem of identity switch during interaction between objects. To resolve the ambiguity between objects that are spatially close to each other, we propose an adaptive model to learn the change of object appearance online. The appearance model of a target $T_i$, noted $\mathbf{m}_i$, is a combination of $d$ sub-models, where $d$ is the feature dimension. Each sub-model contains a set of $n$ observed feature values.

$$\mathbf{m}_i = \{m_i^1(u), m_i^2(u), \ldots, m_i^n(u) | \forall u \in [1, d]\}. \quad (7)$$

The procedure of model construction and update is demonstrated in Fig. 4. In this example, the features are extracted from normalized pixel templates of size $d = w \times h$. The dimension of each feature vector is given by $m_i^k(u) \in \mathbb{R}^6$, as it encodes RGB values in 3 channels, LBP values in 1 channel, as well as gradient magnitudes and angles. To initialize or update this appearance model, each pixel template within the object region is normalized to the size of $w \times h$ (see Fig. 4 (b)). As shown in Fig. 4(c), the foreground mask $q_j$ is used to determine the visible object region. When the observation is occluded, the occluded area is eliminated from $q_j$. The update rate of each sub-model, noted $\alpha_i(u)$, is dynamically controlled by a softmax function, which depends on the distance between newly observed features and values in the past. We define $\alpha_i(u)$ as follows

$$\alpha_i(u) = \left(1 + \exp\left[\min_k \|f_j(u) - m_i^k(u)\| - \tau_f\right]\right)^{-1}, \quad (8)$$

where $f_j(u)$ is the newly observed feature vector of the same dimension as $m_i^k(u)$. The term $\tau_f$ is the maximum distance threshold in the feature space. New features that vary from the past are more likely to be updated, as they reflect the change of appearance that should be learned.

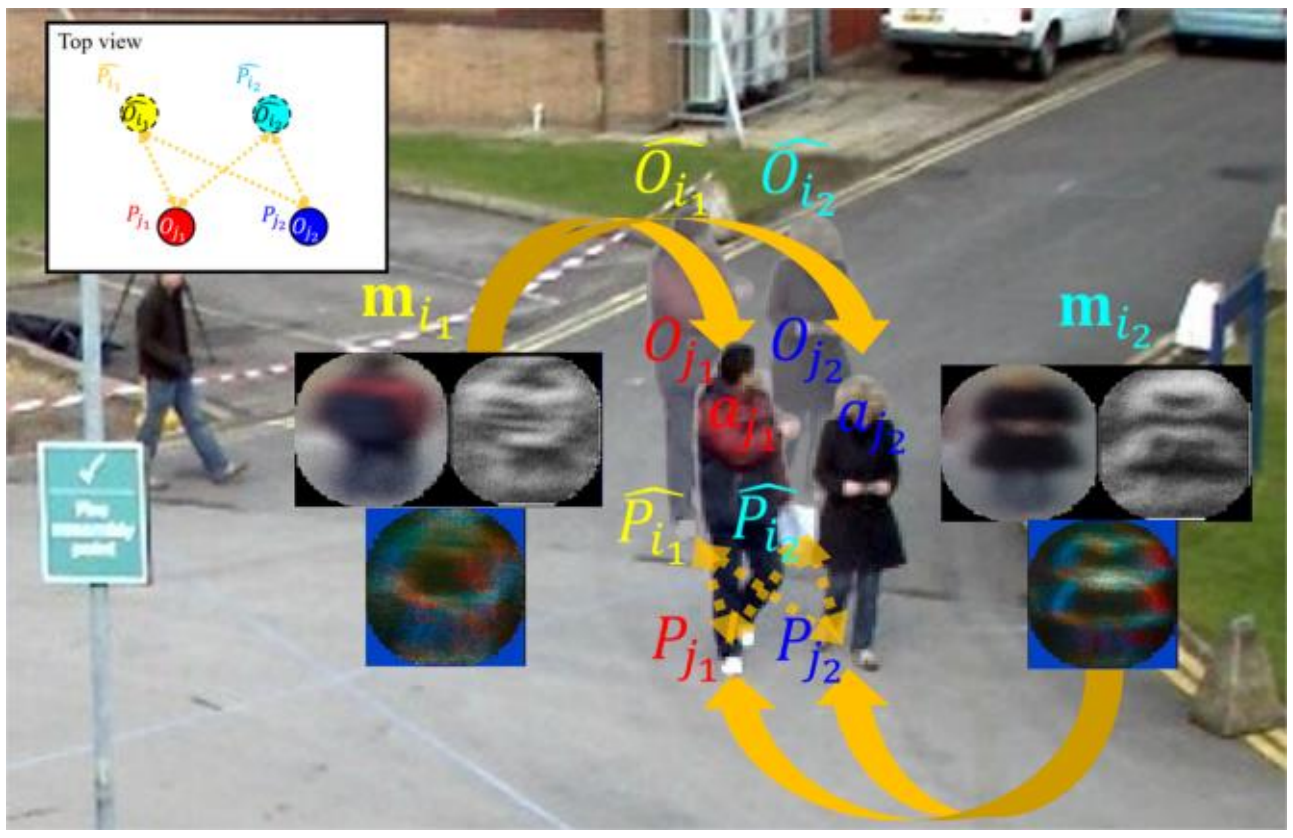

**FIGURE 5. Demonstration of cross-matching for observations grouped with each other, based on 3D geometry information and the proposed adaptive appearance model.**

For pixel-based features like the example in Fig. 4, a Gaussian spatial weighting scheme is also employed to adjust the learning rate as

$$\alpha_i(u) = \frac{\exp\left[-\frac{\|u-u_C\|^2}{2(w^2+h^2)}\right]}{1+\exp\left[\min_k\|f_j(u)-m_i^k(u)\|-\tau_f\right]}, \quad (9)$$

where $u_c$ denotes the center of mass of the visible area within the object region. The spatially weighted learning rates $\alpha_i(u)$ are maximum around the central region, which the body of the object usually occupies, so the sub-models there should be updated more frequently. The learning rate drops as $u$ gets further away from $u_c$. Thus, we can suppress the influence of background area.

The procedure of model update is described as follows. When a candidate observation $O_j$ is associated with $T_i$ in Kalman filtering, the extracted features $f_j$ are used to update the appearance model of $T_i$, i.e., $\mathbf{m}_i$. For each sub-model in $\mathbf{m}_i$, if there are less than $n$ feature vectors stored, the observed feature vector $f_j(u)$ is added into the sub-model by a probability of $\alpha_i(u)$. Otherwise, a random feature vector $m_i^k(u)$ in the sub-model is swapped by $f_j(u)$ with a probability of $\alpha_i(u)$. In Fig. 4(d), each feature component of MOANA is plotted, in which averaged values are displayed.

To measure the appearance affinity using the proposed model, the similarity score between the prediction of $T_i$, noted $\widehat{O_\iota}$, and an observation $O_j$ is given as

$$s(\widehat{O_\iota}, O_j) = \frac{\sum_k\left[\#\left(\left\|f_j(u)-m_i^k(u)\right\|<\tau_f, \forall k\le n\right)\right]}{dn}, \quad (10)$$

where $\#(\cdot)$ returns the number of samples satisfying the given condition. The value of $s(\widehat{O_\iota}, O_j)$ is between 0 and 1, where higher value indicates higher similarity, because more features are matched between the prediction and the observation.

Note that the proposed appearance model is universal, i.e., compatible with all kinds of feature combinations, as long as the feature dimension is fixed. Thus, in the example of Fig. 4, all the pixel templates need to be normalized to $w \times h$. MOANA is also compatible with different measurements of distance in the feature space. Besides, the computation of model update and comparison is always constant, i.e., $O(dn)$. With reasonable setting of algorithmic parameters, the processing speed can be sufficiently fast to support real-time application. Moreover, different from previous approaches, since a set of previously observed feature values is stored and updated in random, MOANA is capable of "memorizing" a relatively long-term history of appearance change, which may cover different viewing angles, object poses and illumination. The proposed method also benefits from the normalized similarity score between 0 and 1, which makes it convenient to set thresholds and compare with each other. On the other hand, common affinity measurements, such as Bhattacharya distance and KL divergence, do not share such property.

### C. CROSS-MATCHING WITH APPEARANCE MODEL

The cross-matching module is enabled when a candidate observation is spatially close to other object(s) but has more than 50% of the object region visible, i.e., in the grouping state. In this case, a predicted target location by Kalman filter may be associated with a wrong observation easily, which leads to identity switch. The problem can be mitigated by comparing the appearance features across grouped objects, i.e., cross-matching, but the effect is limited when the nearby targets share high appearance similarity. Since long-term appearance change is effectively encoded in our proposed appearance model, we can maximally distinguish highly similar objects through cross-matching.

The procedure of cross-matching is demonstrated in Fig. 5. More specifically, for each observation $O_j$ a list of nearby target predictions, noted $l_j$, is kept. If there are more than one prediction in $l_j$, $O_j$ is in the grouping state. In cross-matching, the observation $O_j$ is compared with each element in $l_j$. The computation of similarity score incorporates both 3D geometry information and appearance affinity, defined as

$$s_c(\widehat{O_\iota}, O_j) = s(\widehat{O_\iota}, O_j) \cdot \frac{w_j^D}{\|\widehat{P_\iota} - P_j\|_2}, \widehat{O_\iota} \in l_j, \quad (11)$$

where the subscript $c$ refers to cross-matching. The similarity score in Equ. (10) is divided by the Euclidean distance of 3D foot points, because spatially close objects are more likely to be associated. Similar to Equ. (5), the term $w_j^D$ is added to compensate for the confusion of foot point estimation for distant objects. With the set of computed scores $\{s_c(\widehat{O_\iota}, O_j)\}$ between each pair of observation and prediction, we formulate a bipartite matching problem, which can be effectively solved using the Hungarian algorithm. The detailed pseudocode of the above procedure is provided in Algorithm 1.

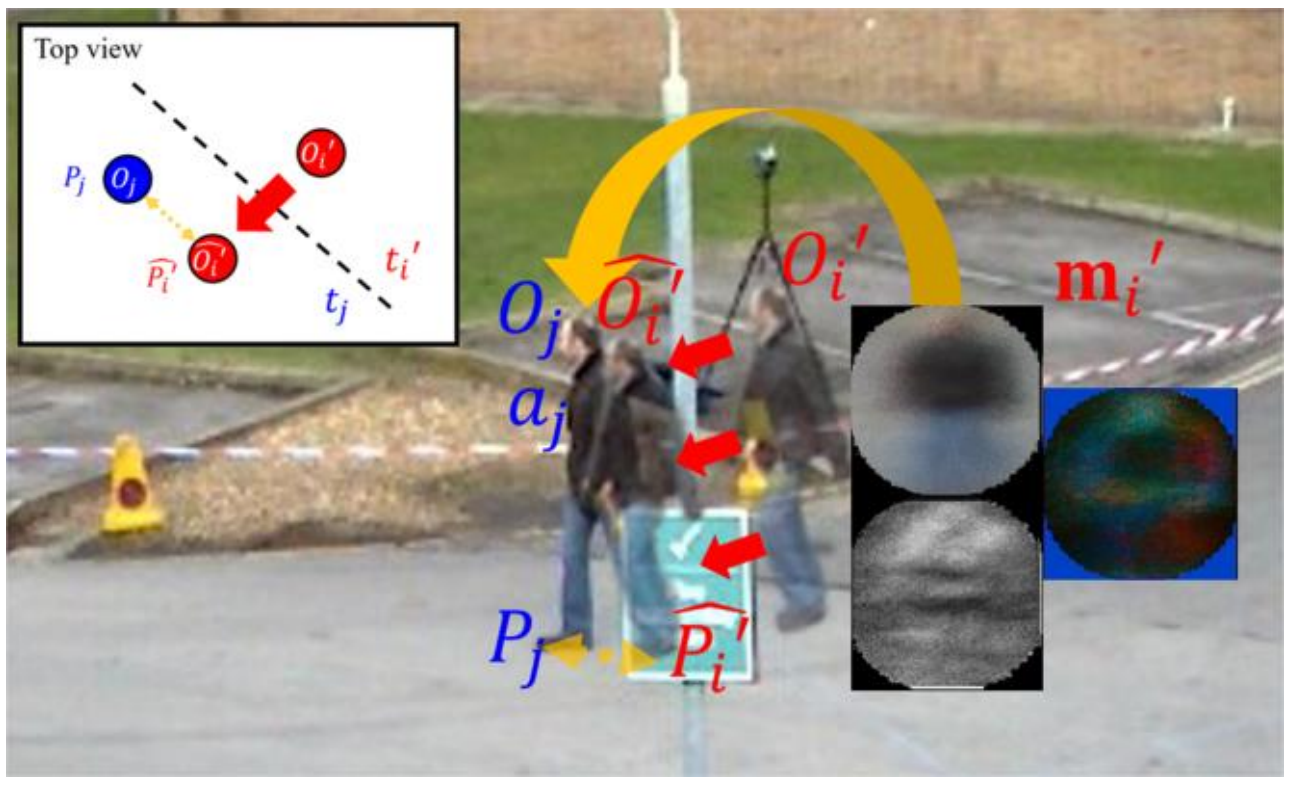

**FIGURE 6.** **Demonstration of re-identification for missing observations, based on 3D geometry information and the proposed adaptive appearance model. A disappeared object, $T_i'$, is shifted to a predicted location, i.e., $\widehat{O_i}'$, and compared with an entering observation $O_j$.**

**Algorithm 1**: Cross-matching based on MOANA

**input**: current video frame, candidate observations in the input frame $\{O_j\}$, prediction of each target from the Kalman filter $\{\widehat{O_i}\}$
**output**: matched pairs of observations and predictions

```
for each O_j in {O_j}
   clear the list of nearby candidate predictions l_j;
   for each Ô_i in {Ô_i}
      if ||P̂_i − P_j||_2 / w_j^D < τ_P or b̂_i overlaps with b_j and
         (visible area of b_j) / (total area of b_j) > 50% do
         push Ô_i into l_j;
      end if
   end for
   if #(l_j) > 1 do
      for each Ô_i in l_j
         compute s_c(Ô_i, O_j) using Equ. (11);
         push s_c(Ô_i, O_j) into {s_c(Ô_i, O_j)};
      end for
   end if
end for
solve the association based on {s_c(Ô_i, O_j)} using
   the Hungarian algorithm;
output all matched pairs of T_i and O_j.
```

## D. RE-IDENTIFICATION WITH APPEARANCE MODEL

When an object observation is under serious occlusion, i.e., the visible area is smaller than 50% or there is no nearby target prediction (false negative), his/her leaving time stamp, location, and appearance model are temporarily stored for re-identification. Since the viewpoint of a target usually changes significantly after serious occlusion, and targets frequently enter and exit the *region of interest* (ROI) in real world, a reliable appearance descriptor that learns long-term appearance variation is key to the success of re-identification.

The procedure of re-identification is demonstrated in Fig. 6. For each entering observation $O_j$ that is not associated with any existing target, it is compared with a list of disappeared targets, noted $T'$. If a missing target is successfully associated with an entering observation, its identity and appearance model is recovered. The similarity score for re-identification is computed as

$$s_r\left(\widehat{O_i}', O_j\right) = \begin{cases} s\left(\widehat{O_i}', O_j\right) \cdot \frac{(t_j - t_i')\cdot w_j^D}{\left\|\widehat{P_i}' - P_j\right\|_2}, & \text{if } \frac{\left\|\widehat{P_i}' - P_j\right\|}{(t_j - t_i')\cdot w_j^D} < \tau_P \\ 0, & \text{otherwise} \end{cases}, \quad (12)$$

in which the subscript $r$ stands for re-identification. $\widehat{O_i}'$ is the Kalman prediction of a missing target at $t_j$. $\widehat{P_i}'$ and $t_i'$ are the predicted 3D location at the current frame and the time stamp that the target disappears respectively. Different from Equ. (11), we have a new term, $(t_j - t_i')$, which calculates the time span in seconds that the target has been missing. The intention is that a target missing for a long time usually leads to higher uncertainty in the prediction of location. Moreover, only the prediction(s) in the neighborhood of the observation are considered as candidate(s) for re-identification. Finally, if the observation is not associated with any missing target or the maximum similarity score is considered too low, i.e., smaller than a threshold noted $\tau_s$, it is identified as a new target. The pseudocode of re-identification is detailed in Algorithm 2.

**Algorithm 2**: Re-identification based on MOANA

**input**: current video frame, entering observations in the input frame $\{O_j\}$, prediction of each disappeared target at the current frame $\{\widehat{O_i}'\}$
**output**: object identities of $\{O_j\}$

```
for each O_j in {O_j}
   for each Ô_i' in {Ô_i'}
      if ||P̂_i' − P_j|| / ((t_j − t_i')·w_j^D) < τ_P do
         compute s_r(Ô_i', O_j) using Equ. (12);
         push s_r(Ô_i, O_j) into {s_r(Ô_i, O_j)};
      end if
   end for
   Ô_i'* ← arg max_{∀Ô_i' ∈ {Ô_i'}} {s_r(Ô_i, O_j)};
   if s_r(Ô_i'*, O_j) > τ_s do
      assign the identity of T_i'* to O_j;
   else
      assign a new identity to O_j
   end if
end for
output all identities of {O_j}.
```

## IV. EXPERIMENTAL RESULTS

Extensive experiments are conducted on the publicly available *MOTChallenge* benchmark [20], which is a collection of

TABLE I
COMPARISON OF THE PROPOSED METHOD WITH THE STATE-OF-THE-ART ON THE MOTCHALLENGE 3D BENCHMARK (TEST SEQUENCES)

| Tracker | Tracking Mode | MOTA (%) | MOTP (%) | MT (%) | ML (%) | FP | FN | ID Sw. | Frag. | Hz |
|---|---|---|---|---|---|---|---|---|---|---|
| MOANA (proposed) | online | **52.7** | 56.3 | 28.4 | 22.0 | 2,226 | **5,551** | **167** | 586 | 19.4 |
| DBN [11] | online | 51.1 | 61.0 | **28.7** | 17.9 | 2,077 | 5,746 | 380 | 418 | 0.1 |
| GPDBN [12] | online | 49.8 | **62.2** | 25.7 | 17.2 | 1,813 | 6,300 | 311 | **386** | 0.1 |
| GustavHX* | online | 42.5 | 56.2 | 25.7 | 15.7 | 2,735 | 6,623 | 302 | 431 | 0.0 |
| MCFPHD [2] | offline | 39.9 | 53.6 | 25.7 | 16.8 | 3,029 | 6,700 | 363 | 529 | 17.7 |
| MCG* | offline | 35.9 | 54.8 | 8.2 | 25.7 | **1,600** | 8,464 | 692 | 1,017 | 0.1 |
| LPSFM [1] | offline | 35.9 | 54.0 | 13.8 | 21.6 | 2,031 | 8,206 | 520 | 601 | 8.4 |
| LP3D [1] | offline | 35.9 | 53.3 | 20.9 | 16.4 | 3,588 | 6,593 | 580 | 659 | **83.5** |
| SVT [3] | offline | 34.2 | 55.8 | 11.2 | 25.4 | 3,057 | 7,454 | 532 | 611 | 1.9 |
| AMIR3D [13] | online | 25.0 | 55.6 | 3.0 | 27.6 | 2,038 | 9,084 | 1,462 | 1,647 | 1.2 |
| KalmanSFM [4] | online | 25.0 | 53.6 | 6.7 | **14.6** | 3,161 | 7,599 | 1,838 | 1,686 | 30.6 |

Bold entries indicate the best results in the corresponding columns.
* GustavHX and MCG are anonymous submissions.

existing and new data for MOT evaluation. It is developed to bring forward the strengths and weaknesses of the state-of-the-art MOT methods. Among all the subsets of this benchmark [20], [30], *MOTChallenge* 2015 3D is the only one dedicated for the evaluation of 3D tracking performance, in which all the videos are taken by static cameras, so that our camera self-calibration scheme can be applied. There are two training sequences, *PETS09-S2L1* and *TUD-Stadtmitte*, with 974 frames and 5,632 ground-truth bounding boxes for 29 targets in total. *AVG-TownCentre* and *PETS09-S2L2* are the two test sequences, which are significantly more complex than the training set, including 886 frames and 16,789 ground truths for 268 targets. The benchmark presents all kinds of evaluation metrics for the performance of MOT [21], [31], such as MOTA, *multiple object tracking precision* (MOTP), *false positives* (FP), *false negatives* (FN), *identity switches* (ID Sw.), *mostly tracked targets* (MT), *mostly lost targets* (ML), *fragments* (Frag.), etc. The two test sequences in the *MOTChallenge* 3D benchmark are included in the *MOTChallenge* 2D benchmark [20] as well, which also allows us to compare with the state-of-art in 2D MOT [32], [13], [33], [34], [35], [36], [37], [38], [39].

The proposed framework has been implemented in C++ with the support of the OpenCV 3 library. It is run on an Intel Core i7-7700K PC with 4 cores, 4.20 GHz processor and 24 GB RAM in the Ubuntu 14.04 environment. After testing different features including pixel templates, histograms, deep learning features and their combinations on the training sequences (to be presented in Section IV-D), we choose to incorporate both RGB and LBP pixel templates in our appearance model for the evaluation on the test sequences. The distance measurement in feature space is given by the Euclidean distance. The minimal color distance threshold and minimal LBP distance threshold, i.e., $\tau_f$, are both empirically set to 30. In all the experimental sequences, the normalized size for feature extraction is empirically set as $w \times h = 64 \times 64$, which is an ideal balance between HD resolution and real-time computation. Due to the relatively short-term appearance of most objects in these sequences, $n$ is set to 3 seconds. In addition, the values of $\tau_P$, $\tau_s$, $\eta_D$ and $c_D$ are empirically chosen to be 2 meters, 0.30, 1/30 and 1 respectively. Moreover, the gaps between re-identified tracklets are linearly interpolated. To conform with the provided ground truth of camera parameters, we compute the transformation from the estimated projection matrix to the actual homography, so that our 3D tracking results can be converted properly for evaluation. The unit used for all 3D measurements is meter.

### A. COMPARISON WITH THE 3D STATE-OF-THE-ART

Currently, there have been 11 submissions on the *MOTChallenge* 2015 3D benchmark, including two anonymous methods. All the experimental results are summarized in Table I. The corresponding qualitative visualization is available in Fig 7. The demo videos can be viewed on the *MOTChallenge* website.[2] Note that the noisy detection sets provided by the benchmark are used as input to our algorithm. To be fair with other methods in comparison, we do not apply foreground segmentation in our appearance model. Thus, each object mask in the proposed appearance model is defined by a maximum ellipse, as shown in Fig. 4(c).

MOANA is currently ranked on top in terms of the two most significant metrics, MOTA and ID Sw. As shown in Fig. 7 and the online demo videos, our predicted trajectories and localization of targets are all relatively more accurate, whereas other methods missed a few more targets and introduced more false positives. The promising performance mainly benefits from the proposed appearance adaptation scheme that maintains robustness against occlusion and appearance similarity among nearby targets. This is proven by the fact that our ID Sw. on this challenging benchmark is reduced by over 46% compared with the former leader [12]. This also explains why MOANA enjoys a relatively high MT score. However, a drawback of the interpolation scheme is that the number of fragments will increase caused by growing FP, as some objects may not walk linearly under serious occlusion. Nonetheless, the negative influence on our overall performance can be neglected.

[2]Available at https://motchallenge.net/vis/PETS09-S2L2/MOANA and https://motchallenge.net/vis/AVG-TownCentre/MOANA

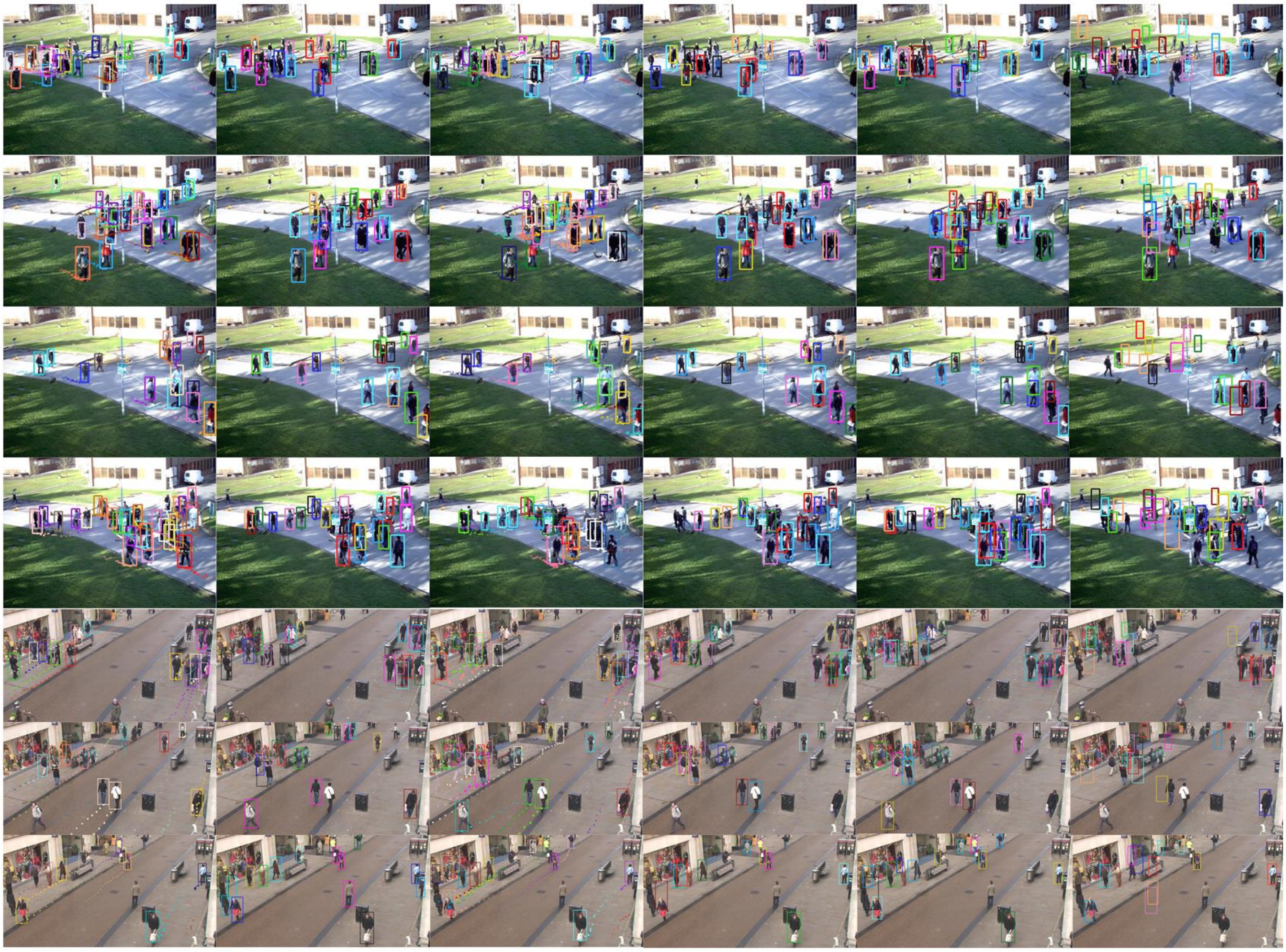

**FIGURE 7. Qualitative comparison on the test sequences of the *MOTChallenge* 3D benchmark, which can be better visualized through demo videos on the *MOTChallenge* website. *First row*: Frame #91 of *PETS09-S2L2*. *Second row*: Frame #222 of *PETS09-S2L2*. *Third row*: Frame #409 of *PETS09-S2L2*. *Fourth row*: Frame #128 of *AVG-TownCentre*. *Fifth row*: Frame #189 of *AVG-TownCentre*. *Sixth row*: Frame #441 of *AVG-TownCentre*. *First column*: MOANA. *Second column*: DBN [11]. *Third column*: MCFPHD [2]. *Fourth column*: LPSFM [1]. *Fifth column*: LP3D [1]. *Sixth column*: KalmanSFM [4].**

Among other state-of-the-art in comparison, DBN [11] and GPDBN [12] gain the second and third places in the ranking, which both apply a Bayesian filtering approach, named *dynamic bayes network* (DBN), for state prediction. The changing appearance of each target is learned online based on a *random forest* formulation. Because of similar improvement in appearance modeling, their MOTA score is only inferior to ours by margin, but they have better performance on MOTP, MT and ML. MCFPHD [2] utilizes PHD filter for instantaneous multi-target state estimation. The decisions on target trajectories are made offline. Likewise, LPSFM and LP3D [1], the baselines on this benchmark, make use of linear programming and social force model for data association in 3D, which are also offline methods. The works [2], [1] focus on modeling the motion patterns but do not incorporate any appearance model in their formulation, which explains why their performance is inferior to MOANA and DBN-based methods. SVT [3] explores the use of spatio-temporal hypergraph to encode 3D constraints and appearance information, however, their appearance model is based on color histogram from single image, which can be easily affected by appearance change. AMIR3D [13] is another new method that exploits RNNs to jointly reason multiple cues for tracking, including appearance similarity. The recently observed deep learning features are kept in a feature vector, but the history beyond a temporal window is discarded absolutely, which is less reliable than our strategy based on random update. Furthermore, the deep learning features are similar among objects within the same class, which may not perform well for discriminative appearance modeling. Finally, the

TABLE II
COMPARISON OF THE PROPOSED METHOD WITH THE STATE-OF-THE-ART ON THE MOTCHALLENGE 2D BENCHMARK (AVG-TOWNCENTRE)

| Tracker | Tracking Mode | MOTA (%) | MOTP (%) | MT (%) | ML (%) | FP | FN | ID Sw. | Frag. |
|---|---|---|---|---|---|---|---|---|---|
| MOANA (proposed) | online | **46.1** | 55.1 | 26.1 | 24.8 | 773 | 3,020 | 60 | **200** |
| AP_HWDPL_p [32] | online | 28.4 | 66.9 | 4.0 | 27.9 | 941 | 4,005 | 169 | 412 |
| AMIR15 [13] | online | 36.2 | 69.5 | 26.1 | **17.7** | 1,448 | **2,882** | 234 | 389 |
| JointMC [33] | offline | 43.1 | 69.8 | **29.2** | 32.3 | 922 | 3,116 | **28** | 213 |
| HybridDAT [34] | online | 29.2 | 69 | 9.3 | 43.4 | **532** | 4,465 | 61 | 246 |
| AM [35] | online | 37.5 | 68.1 | 14.2 | 30. | 645 | 3,742 | 79 | 332 |
| TSMLCDEnew [36] | offline | 33.9 | 68.9 | 20.4 | 31.0 | 997 | 3,604 | 126 | 274 |
| QuadMOT [37] | offline | 30.8 | 69.8 | 18.1 | 31.4 | 1,191 | 3,643 | 111 | 409 |
| NOMT [38] | offline | 31.6 | **70.1** | 11.1 | 36.3 | 681 | 4,060 | 146 | 233 |
| DCCRF [39] | online | 32.3 | 68.9 | 12.4 | 29.2 | 777 | 3,831 | 229 | 229 |

Bold entries indicate the best results in the corresponding columns.

TABLE III
COMPARISON OF THE PROPOSED METHOD WITH THE STATE-OF-THE-ART ON THE MOTCHALLENGE 2D BENCHMARK (PETS09-S2L2)

| Tracker | Tracking Mode | MOTA (%) | MOTP (%) | MT (%) | ML (%) | FP | FN | ID Sw. | Frag. |
|---|---|---|---|---|---|---|---|---|---|
| MOANA (proposed) | online | **57.6** | 57.0 | **40.5** | 7.1 | 1,453 | **2,531** | **107** | 386 |
| AP_HWDPL_p [32] | online | 38.9 | 70.8 | 2.4 | 9.5 | **552** | 5,164 | 179 | 328 |
| AMIR15 [13] | online | 47.0 | 70.5 | 11.9 | 9.5 | 616 | 4,236 | 254 | 397 |
| JointMC [33] | offline | 56.0 | 71.4 | 23.8 | **4.8** | 942 | 3,162 | 142 | 220 |
| HybridDAT [34] | online | 47.7 | 69.3 | 11.9 | 9.5 | 616 | 4,236 | 254 | 349 |
| AM [35] | online | 47.7 | 69.2 | 16.7 | 14.3 | 718 | 4,206 | 115 | 356 |
| TSMLCDEnew [36] | offline | 51.5 | 70.6 | 14.3 | 9.5 | 905 | 3,602 | 165 | **198** |
| QuadMOT [37] | offline | 49.0 | **72.6** | 16.7 | 7.1 | 686 | 3,947 | 285 | 380 |
| NOMT [38] | offline | 53.4 | 70.5 | 14.3 | 9.5 | 884 | 3,465 | 142 | 208 |
| DCCRF [39] | online | 45.6 | 72.4 | 9.5 | 9.5 | 664 | 4,335 | 245 | 245 |

Bold entries indicate the best results in the corresponding columns.

unsatisfactory performance of KalmanSFM [4] is also caused by the relatively simple appearance descriptor, which is a raw pixel template that is sensitive to noise. As shown in Fig. 7, several false positives are introduced by their approach.

It is also interesting to study the performance of the state-of-the-art in computation efficiency. With CPU power only, MOANA is able to achieve real-time performance with an average processing speed of 19.4 frames per second on all the test sequences. Even though there are many cases of occlusion and grouping of targets in these sequences that require massive comparison based on the adaptive appearance models, our runtime is not seriously degraded, because our strategy of similarity measurement based on feature distance and spatial weighting is relatively efficient. On the contrary, the computation speed of DBN-based methods using *random forest* is much slower and far from real time. The offline methods [4], [1], [2] are all relatively much faster, because they either do not use appearance model or only use simple representation for their purpose. It is impressive that MOANA can gain a comparable processing speed with them while capable of running online.

### B. COMPARISON WITH THE 2D STATE-OF-THE-ART

Because of the application of camera self-calibration [28], [29], the provided camera matrices in the *MOTChallenge* 3D benchmark are not adopted in our 3D MOT computation, but are only considered for evaluation. Therefore, our algorithm actually only leverages 2D information for 3D MOT. Our superior performance over the state-of-the-art in 3D MOT verifies the effectiveness of our self-calibration scheme.

The two test sequences in the *MOTChallenge* 3D benchmark, *AVG-TownCentre* and *PETS09-S2L2*, are also included in the 2D benchmark. The proposed method is also compared with some of the state-of-the-art 2D MOT methods on these sequences. The experimental results are respectively presented in Table II and Table III. Note that because a different evaluation scheme for object localization is adopted in the 2D MOT dataset, the MOTP scores of all the methods are generally higher than those in the 3D MOT benchmark. Nonetheless, our proposed algorithm still demonstrates significant advantage in MOTA and ID Sw. against them.

TABLE IV
COMPARISON OF VARIANTS OF THE MOANA ALGORITHM ON THE MOTCHALLENGE 3D BENCHMARK (TRAINING SEQUENCES)

| Tracker | MOTA (%) | MOTP (%) | MT (%) | ML (%) | ID Sw. |
|---|---|---|---|---|---|
| MOANA (proposed) | **81.5** | 70.8 | **89.7** | **0.0** | **3** |
| MOANA w/o cross-matching | 68.1 | 68.7 | 51.7 | 24.1 | 47 |
| MOANA w/o re-identification | 64.1 | 69.6 | 62.1 | 10.3 | 34 |
| RPT w/ cross-matching & re-id | 64.3 | 70.0 | 51.7 | 27.6 | 32 |
| MOANA w/o adaptive update | 77.7 | 69.7 | 82.8 | 6.9 | 12 |
| MOANA w/o spatial weighting | 80.5 | 70.5 | 86.2 | 6.9 | 7 |
| Baseline | 77.5 | **72.0** | 79.3 | 3.4 | 37 |

Bold entries indicate the best results in the corresponding columns.

### C. ABLATION STUDY

We conduct more experiments with variants of our proposed method on the training sequences of the *MOTChallenge* 3D benchmark, as the test sequences are not allowed for self-comparison. The results are summarized in Table IV.

The proposed adaptive appearance models are applied to two data association schemes, namely cross-matching and re-identification, respectively. As we have expected, when both schemes are taken into account, we achieve the best performance in the majority of measurements. When cross-matching is not considered, a large number of identity switches occur, because of spatial ambiguity among adjacent targets. On the other hand, when re-identification is not adopted, the identities of temporarily occluded targets cannot be recovered, which also leads to inferior performance. We also compare with the appearance model from the *raw pixel template* (RPT), i.e., the latest available instance from a single frame (see Fig. 4(b)). The main difference is that a long-term history of appearance change is learned by our proposed appearance model. As can be seen from the comparison, RPT with the proposed cross-matching and re-identification schemes fail to recover most of the identity switches. Furthermore, we evaluate the proposed adaptive update of learning rates, i.e., Equ. (8). The experimental results prove the effectiveness of adaptive learning in our formulation, as more diverse feature values are kept in our appearance model. Then, to validate the proposed Gaussian spatial weighting scheme for pixel-based appearance modeling, i.e., Equ. (9), we also compare to model update without spatial weighting. As shown in Table IV, our proposed scheme boosts the performance, as the background area is suppressed in feature extraction. Finally, MOANA also demonstrates major improvement over the baseline, especially in the reduction of identity switches.

### D. COMPARISON OF FEATURE COMBINATION

In this subsection, we explore the effectiveness and computation efficiency of different features and their combinations for the proposed appearance modeling scheme. Experiments are conducted on the training set of the *MOTChallenge* 3D benchmark. The experimental results are presented in Table V. Note that the CNN features are extracted from a GoogLeNet [40] pre-trained on the COCO benchmark [41], with a feature dimension of 1,024. For the histogram-based features, all the feature channels have 8 bins each. The Gaussian spatial weighting scheme is not applied to the extraction of CNN features, but it is employed for the pixel-based description and histogram construction. For all the feature comparison, we adopt the Euclidean distance.

TABLE V
COMPARISON OF FEATURE COMBINATIONS FOR THE MOANA ALGORITHM ON THE MOTCHALLENGE 3D BENCHMARK (TRAINING SEQUENCES)

| Tracker | MOTA (%) | MOTP (%) | MT (%) | ML (%) | ID Sw. | Hz |
|---|---|---|---|---|---|---|
| Pix.: RGB + LBP + Grad. | 81.7 | 70.5 | **89.7** | **0.0** | 3 | 28.5 |
| Pix.: RGB + LBP | 81.5 | **70.8** | **89.7** | **0.0** | 3 | 29.7 |
| Pix.: RGB + Grad. | 80.4 | **70.8** | 82.8 | **0.0** | 5 | 30.6 |
| Pix.: LBP + Grad. | 76.3 | 70.0 | 72.4 | 6.9 | 15 | 31.5 |
| Pix.: RGB | 80.2 | 70.5 | 82.8 | **0.0** | 3 | 31.8 |
| Pix.: LBP | 75.7 | 70.2 | 75.9 | 6.9 | 19 | 36.3 |
| Pix.: Grad. | 66.5 | 69.7 | 58.6 | 10.3 | 31 | 32.9 |
| Hist.: RGB + LBP + Grad. | 72.6 | 69.9 | 79.3 | 20.7 | 10 | 36.1 |
| Hist.: RGB + LBP | 72.5 | 69.2 | 75.9 | 13.8 | 15 | 37.7 |
| Hist.: RGB + Grad. | 71.0 | 69.3 | 65.5 | 24.1 | 11 | 38.9 |
| Hist.: LBP + Grad. | 65.2 | 69.4 | 62.1 | 27.6 | 21 | 40.9 |
| Hist.: RGB | 70.2 | 70.3 | 75.9 | 17.2 | 14 | 40.7 |
| Hist.: LBP | 63.0 | 69.1 | 58.6 | 31.0 | 23 | **43.2** |
| Hist.: Grad. | 56.5 | 69.2 | 51.7 | 37.9 | 40 | 41.1 |
| CNN | **82.6** | 70.3 | 86.2 | 0.0 | **1** | 1.6 |

Bold entries indicate the best results in the corresponding columns.

The CNN features and the combination of all pixel-based features, i.e., RGB, LBP and gradient, achieve the best overall performance on the major evaluation metrics. The deep learning features are trained to classify objects with millions of samples, so they lead to higher accuracy in data association, but the feature extraction without GPU is time-consuming. The pixel-based methods demonstrate higher accuracy compared to histogram-based ones, because the spatial feature distribution is explicitly encoded in the pixel templates. We can also learn that the RGB color component contains the richest information in appearance description, as all combinations with the RGB feature generally perform better than others. Finally, the combination of RGB and LBP in pixel templates is chosen for the experiments on the test sequences, because of its robust performance and relatively lower computation requirement. Note that because the crowd of human targets is denser in the test sequences, which requires more computation in cross-matching and re-identification, so the general runtime is slower than the training sequences.

As mentioned in Section III-B, MOANA can be easily receptive to incorporating other useful features or their combination with fixed dimension. Thus, the robustness of the proposed model can be further improved through the combination with more discriminant image features, such as [42], [43], [44], [45], [46], [47].

## V. CONCLUSIONS

Multi-target tracking has been a challenging task, especially because of identity switches caused by occlusion, spatial ambiguity and similar appearance among nearby targets. In this paper, we propose an adaptive appearance modeling scheme to support robust MOT. Different from previous works in the development of discriminative appearance features, our extracted feature vectors are saved in an explicit form and adaptively updated online. The proposed method is robust against appearance change due to different illumination, poses and viewing perspectives. Based on the adaptive appearance model, we design cross-matching and re-identification schemes to mitigate identity switch when objects interact with others. Besides, 3D geometry information is effectively incorporated into our formulation of data association. Experimental results on the *MOTChallenge* benchmark datasets show our superior performance in robustness and efficiency compared with the state-of-the-art. In the future, we plan to extend our work to moving cameras with the assistance of visual odometry or visual SLAM. We will also leverage saliency detection techniques [48], [49], [50], [51] to improve edge-based appearance modeling. Last but not least, MOT can be largely benefited from improved object detectors such as [52], [53].